\documentclass[10pt,twocolumn,letterpaper]{article}

\usepackage{cvpr}              

\usepackage{graphicx}
\usepackage{amsmath}
\usepackage{amssymb}
\usepackage{booktabs}
\usepackage[accsupp]{axessibility}  
\usepackage{multirow}
\usepackage{ulem}
\def\etal{\emph{et al}.}
\def\eg{\emph{e.g}.}
\def\ie{\emph{i.e}.}

\newcommand{\wh}[1]{\textcolor{black}{{#1}}}

\newcommand{\YL}[1]{{\color{black}#1}}

\usepackage[pagebackref,breaklinks,colorlinks]{hyperref}

\usepackage[capitalize]{cleveref}
\crefname{section}{Sec.}{Secs.}
\Crefname{section}{Section}{Sections}
\Crefname{table}{Table}{Tables}
\crefname{table}{Tab.}{Tabs.}

\def\cvprPaperID{7792} 
\def\confName{CVPR}
\def\confYear{2023}

\begin{document}

\title{Crowd3D: Towards Hundreds of People Reconstruction from a Single Image}


\author{
    Hao Wen$^{1, \dagger}$, Jing Huang$^{1, \dagger}$, Huili Cui$^{1}$, Haozhe Lin$^{2}$, Yu-Kun Lai$^{3}$, Lu Fang$^{2}$, Kun Li$^{1,*}$\\
    $^{1}$Tianjin University, China \enspace  $^{2}$Tsinghua University, China \enspace $^{3}$Cardiff University, United Kingdom\\
    {\tt\small \{wenhao, hj00, huilicui\_1, lik\}@tju.edu.cn,
    \{linhz, fanglu\}@tsinghua.edu.cn,}\\
    {\tt\small LaiY4@cardiff.ac.uk}
}

\twocolumn[{
\renewcommand\twocolumn[1][]{#1}
\maketitle
\begin{center}
    \captionsetup{type=figure}
    \includegraphics[width=0.95\textwidth]{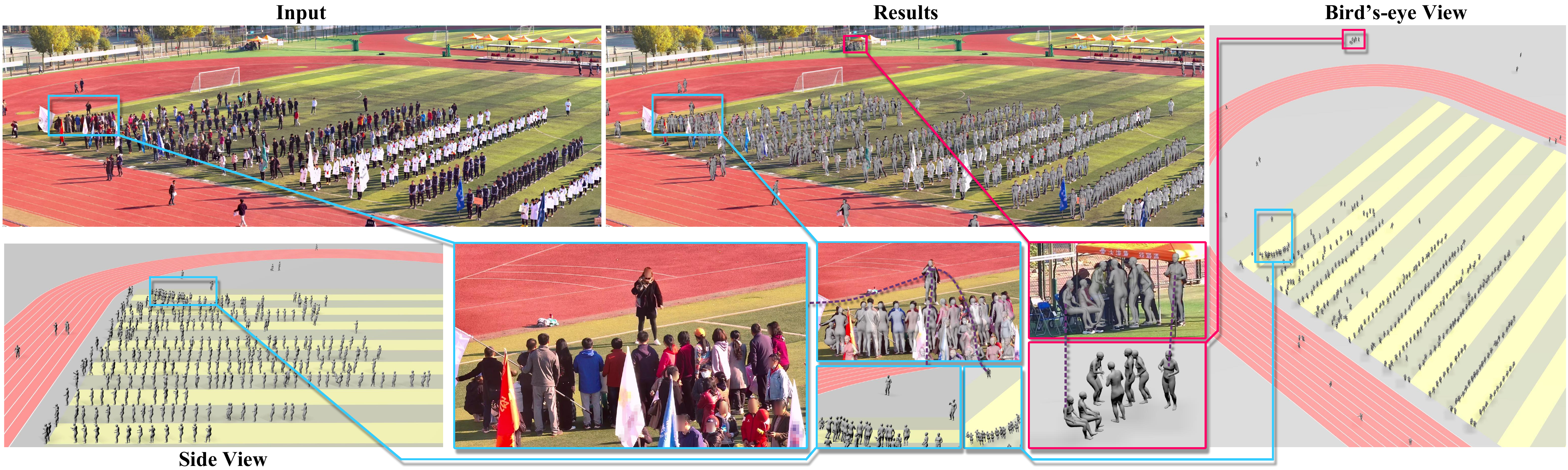}
    \captionof{figure}{Given a single large-scene image with hundreds of people, our method can reconstruct 3D poses, shapes and locations of these people in a global camera space with coherency with the scene. Please zoom in for more details. }
    \label{fig:show}
\end{center}
}]

\begin{abstract}
\vspace{-0.2cm}
    
    \let\thefootnote\relax\footnotetext{$\dagger$ Equal contribution.}
    \let\thefootnote\relax\footnotetext{* Corresponding author.}
   Image-based multi-person reconstruction in wide-field large scenes is critical for crowd analysis and security alert.
   However, existing methods cannot deal with large scenes containing hundreds of people, which encounter the challenges of large number of people, large variations in human scale, and complex spatial distribution.
   In this paper, we propose Crowd3D, the first framework to reconstruct the 3D poses, shapes and locations of hundreds of people with global consistency from a single large-scene image. 
   The core of our approach is \YL{to convert} the problem of complex crowd localization into pixel localization with the help of our newly defined concept, Human-scene Virtual Interaction Point (HVIP).
   To reconstruct the crowd with global consistency, we propose a progressive reconstruction network based on HVIP by pre-estimating a scene-level camera and a ground plane.
   To deal with a large number of persons and various human sizes, we also design an adaptive human-centric cropping scheme. 
   Besides, we contribute a benchmark dataset, LargeCrowd, for crowd reconstruction in a large scene.
   Experimental results
   demonstrate the effectiveness of the proposed method.
   \wh{The code and the dataset are available at \url{http://cic.tju.edu.cn/faculty/likun/projects/Crowd3D}. }
   
\end{abstract}

\vspace{-0.2cm} 
\section{Introduction}\label{sec:intro}

3D pose, shape and location reconstruction for hundreds of people in a large scene will help with modeling crowd behavior for simulation and security monitoring. However, no existing methods can achieve this with global consistency. In this paper, we aim to reconstruct the 3D poses, shapes and locations of hundreds of people in the global camera space from a single large-scene image, as shown in Fig.~\ref{fig:show}. 

Although monocular human pose and shape estimation \cite{HMR, zhang2021pymaf, tian2022recovering} has been extensively explored over the past years, estimating global space locations together with human poses and shapes for multiple people from a single image is still a difficult problem due to the depth ambiguity. Existing methods \cite{CRMH, BEV} reconstruct 3D poses, shapes and relative positions of the reconstructed human meshes by assuming a constant focal length. But the 
\YL{methods are limited to}
small scenes with a common FoV (Field of View). These methods cannot regress the people from a whole large-scene image \cite{PANDA} due to the relatively small and varying human scales in comparison to the image size. 
Even with an image cropping strategy, these methods cannot obtain consistent reconstructions in the global camera space due to independent inference from the cropped images.
Besides, existing methods hardly consider the coherence of the reconstructed people with the outdoor scene, especially with the ground, since the ground is a common and significant element of outdoor scenes. 
Taking a usual urban scene as an example, these methods may include wrong positions and rotations so that the reconstructed people 
\YL{do not appear} to be standing or walking on the ground.

In general, there are three challenges in reconstructing hundreds of people with global consistency from a single large-scene image:
1) there are \YL{a large number} of people with relatively small and highly 
\YL{varying}
2D scales;
2) due to the depth ambiguity from a single view, it is difficult to directly estimate absolute 3D positions and 3D poses of people in the large scene;
3) there is no large-scene image datasets with hundreds of people for supervising crowd reconstruction in large scenes.


In this paper, to address these challenges, we propose \emph{Crowd3D}, the first framework for crowd reconstruction from a single large-scene image.
To deal with the large number of people and various human scales, we propose an adaptive human-centric cropping scheme for a consistent scale proportion of people among different cropped images by leveraging the observation of pyramid-like changes in the scales of people in large-scene images.
To ensure the globally consistent spatial locations and coherence with the scene, we propose a progressive ground-guided reconstruction network Crowd3DNet to reconstruct globally consistent human body meshes from the cropped images by pre-estimating a global scene-level camera and a ground plane.
To alleviate the ambiguity brought \YL{in} by directly estimating absolute 3D locations from a single image,
we present a novel concept called \emph{Human-scene Virtual Interaction Point (HVIP)} for effectively converting the 3D crowd spatial localization problem into a progressive 2D pixel localization problem with intermediate supervisions. Benefiting from HVIP, our model can reconstruct the people with various poses including non-standing. 

We also construct \emph{LargeCrowd}, a benchmark dataset with over 100K labeled humans (2D bounding boxes, 2D keypoints, 3D ground plane and  HVIPs) in 733 gigapixel images (19200×6480) of 9 different scenes. To our best knowledge, this is the first large-scene crowd dataset, which enables the training and evaluation on large-scene images with hundreds of people.
Experimental results demonstrate that our method achieves globally consistent crowd reconstruction in a large scene.
Fig.~\ref{fig:show} gives an example.

To summarize, our main contributions include:
\vspace{-0.2cm} 
\begin{itemize}
\item [1)] 
We propose Crowd3D, 
a multi-person 3D pose, shape and location estimation framework for large-scale scenes with hundreds of people. 
We design an adaptive human-centric cropping scheme and a joint local and global strategy to achieve the globally consistent reconstruction.
\vspace{-0.2cm} 
\item [2)]
We propose a progressive reconstruction network with \YL{the} newly defined HVIP, to alleviate the depth ambiguity and obtain global reconstructions in harmony with the scene. 
\vspace{-0.2cm} 
\item [3)]
We contribute \emph{LargeCrowd}, a benchmark dataset with over 100K labeled crowded people in 733 gigapixel large-scene images (19200×6480), 
which are valuable for the training and evaluation of crowd reconstruction and spatial reasoning in large scenes. 
\end{itemize}

\section{Related Work}
\vspace{-0.1cm} 
\label{RelatedWork}
\textbf{Multi-person 3D Pose Estimation.} These methods can be divided into top-down \cite{3DMPPE,PANDANet,Lcr-net++,hmor} or bottom-up \cite{smap,mehta2020xnect,MuCo3DHP,fabbri2020compressed,kundu2020unsupervised,PAF} paradigms. The top-down methods first detect the people and then estimate the 3D pose of each person separately. 
Moon \etal~\cite{3DMPPE} estimate root location and root-relative pose separately after detecting the persons. They regard the area of 2D bounding box as a prior and adopt a neural network to learn a correction factor. HMOR \cite{hmor} divides human relations into three levels and formulates pair-wise ordinal relations in each level. Different from the top-down paradigm, the bottom-up methods directly detect all the joints and group them. However, most methods either optimize the translation in a post-processing way \cite{mehta2020xnect} or ignore the root localization. Inspired by monocular depth estimation methods, SMAP \cite{smap} utilizes \YL{a} deep convolutional neural network (CNN) to estimate a normalized root depth map and part relative-depth maps. The final root map is recovered with the given focal length and hence the camera parameters need to be known to obtain the absolute positions.

All the above methods only estimate 3D poses in the form of skeletons while missing shape information that is important for many applications, such as interpenetration reasoning to avoid impossible poses, person re-identification and crowd analysis. 

\begin{figure*}[htb]
    \centering
    \includegraphics[width=0.98\linewidth]{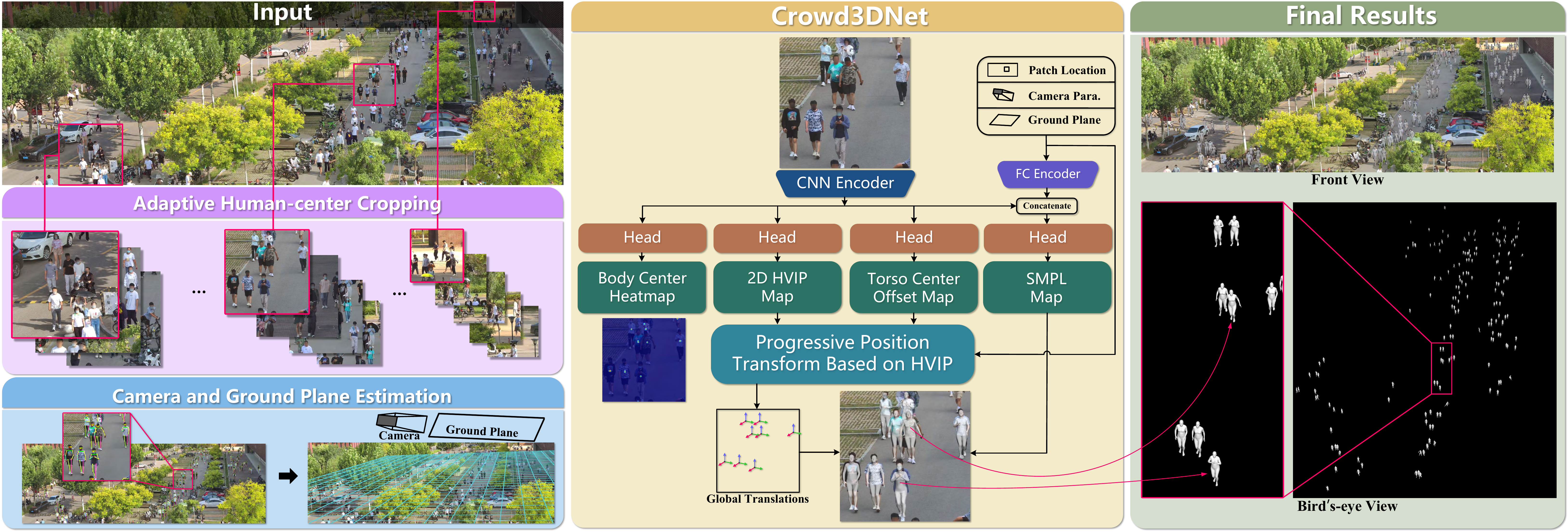}
    \vspace{-0.2cm}
    \caption{Overview of Crowd3D framework. First, Crowd3D adopts an adaptive human-centric cropping scheme to crop the large-scene image into patches with hierarchical sizes for more appropriate scales of people. Then, Crowd3D estimates the scene-camera intrinsics and ground plane equation with human pose priors. Finally, Crowd3DNet takes the cropped image, the patch location, the estimated camera and ground parameters as inputs and outputs the crowd \YL{reconstruction} with consistent spatial locations in the global camera space.}
    \label{fig:overview}
    \vspace{-0.2cm}
\end{figure*}

\textbf{Multi-person 3D Pose and Shape Estimation.} Parametric human body models, \eg, SMPL \cite{smpl}, have been widely adopted to represent the 3D pose and shape of a person. 
Single-person 3D pose and shape estimation has been achieved with tremendous progress \cite{KEEPitSMPL,HMR,vibe,spec,kolotouros2019learning,moon2020pose2pose,pavlakos2019texturepose,sun2019human,zeng20203d,zhang2021pymaf}, while multi-person 3D pose and shape estimation still faces many challenges. 

Some methods adopt a two-stage framework by utilizing a single-person reconstruction method for each detected person. 3DCrowdNet \cite{3DCrowdNet} leverages 2D poses to distinguish different people and uses a joint-based regressor to estimate human model parameters. This kind of approaches focuses more on the accuracy of pose and shape but ignores 3D spatial locations of the people which are important for holistic understanding of the scene.
To get coherent reconstruction results, Jiang \etal~\cite{CRMH} propose CRMH, an R-CNN-based architecture, to detect all the people in an image and estimate their SMPL parameters by using an interpenetration loss and a depth ordering-aware loss in training.
This method calculates human depths based on the assumption that people are consistent in height, which will estimate excessive depths for short individuals like kids. 
To solve the inherent body size and depth ambiguity problem, Ugrinovic \etal~\cite{bodysize} propose a multi-stage optimization-based method to optimize the 3D translations and scales of body meshes estimated by CRMH \cite{CRMH}. Different from multi-stage methods with computation redundancy, BMP \cite{BMP} is a single-stage solution for multi-person mesh regression, which correlates the depth of a person with the features of different scales.  In ROMP \cite{ROMP}, the mesh and location information can be obtained in combination with the camera map and SMPL map according to the center map. However, this method is based on the assumption of weak perspective projection and can only reason about the 2D locations of people in the image plane. It uses an approximation method to obtain depth ordering. 
To address this, BEV \cite{BEV} uses Bird's-Eye-View representation to simultaneously reason about body centers in image and in depth. As mentioned by itself \cite{BEV}, BEV is not trained or designed to deal with large ``crowds'' (\eg, \YL{100s} of people) with a constant focal length assumption. In general, all the above methods can only get relative depths rather than absolute 3D positions, and they cannot be applied directly to large scenes. 

\textbf{Multi-person Datasets.}
Multi-person datasets can be collected in  indoor controlled environments or in outdoor scenes.
Datasets such as \emph{Panoptic} \cite{Panoptic} build multi-view capture systems to obtain relatively high accurate ground-truths. For outdoor scenes, some datasets enable in-the-wild 3D capture from videos with IMUs \cite{3DPW} or from videos in which humans have to stay still \cite{smply}, while other datasets \cite{MPI} annotate in-the-wild images in 2D only. Besides, datasets such as \emph{Agora} \cite{Agora} and \emph{MuCo-3DHP} \cite{MuCo3DHP} generate synthetic images including 3D people and background images or 3D background scenes. 
However, all the above datasets only contain a few people in small scenes.

In this paper, We propose the first work to reconstruct hundreds of people in a large scene with global consistency from a single RGB image.
We also contribute a benchmark dataset, \emph{LargeCrowd}, for the training and evaluation of crowd reconstruction in large scenes. 

\vspace{-0.1cm} 
\section{Method}
\vspace{-0.1cm} 
\label{method}

Our work aims to recover a globally coherent reconstruction of crowd from a single large-scene image with hundreds of persons.
Fig.~\ref{fig:overview} shows the framework of our method.
The highlight of our method is that we design progressive position transform with our newly defined concept HVIP to establish a mapping between local image points and global spatial positions.
Our method consists of three main steps: 
1) we adopt an adaptive human-centric cropping scheme (Sec. \ref{subsec:cropping}) to crop the large-scene image into patches with hierarchical sizes which
ensures that people in different cropped images have appropriate scales;
2) we estimate the camera intrinsics and ground plane equation (Sec. \ref{subsec:groundEstimation}) of the scene with human pose priors for subsequent inference; 
3) taking the cropped images, ground plane and camera parameters as inputs, we design the Crowd3DNet (Sec. \ref{subsec:crowd3dnet}) with the progressive position transform based on HVIP (Sec. \ref{subsubsec:ground-guided}) to directly estimate the human meshes in the large-scene camera coordinate system.


\vspace{-0.1cm} 
\subsection{Adaptive Human-centric Cropping}
\vspace{-0.1cm} 
\label{subsec:cropping}
Instead of using uniform cropping \cite{ferreira2020s,li2022region} that cannot deal with people of various \YL{image} sizes,
we propose an adaptive human-centric cropping strategy to ensure that the height ratio between people and the corresponding cropped image is as consistent as possible among different cropped images.
It is crucial for accurate and reasonable estimation.
Inspired by the observation that 
human heights hierarchically vary like a pyramid in the vertical direction of large-scene image,
the sizes of the cropped images \YL{should also} conform to a similar hierarchical change.
Heuristically, 
we use \YL{a} geometric sequence to simulate the hierarchical change, which is simple but effective.
Define the heights of the persons at the top and the bottom of the large-scene image as $h_{\text{t}}$ and $h_{\text{b}}$, respectively. The upper and lower bounds of the image area \YL{to be processed} are defined as $b_u$ and $b_l$.
\YL{Considering} non-overlapping square blocks in \YL{the} vertical direction of image, \YL{we} represent the sizes of blocks from top to bottom \YL{as} $\{c_i\}_{i=1}^{n}$. When we set the height of people in \YL{a} block to be half of the block size and make $\{c_i\}_{i=1}^{n}$ comply with the rule of geometric sequence, we have $c_1=2 \times h_{\text{t}}$, \wh{$c_i=c_1 \times q^{i-1}$} and $\sum_{i=1}^{n}c_i=b_l-b_u$, where $q$ is the proportionality coefficient.
This cropping problem is formulated as:
\begin{equation}
    \label{eq:cropping}
     \mathop{\arg\min}\limits_{n, q}|c_n-2\times h_{\text{b}}|.
\end{equation}
\wh{To ensure each person can appear completely in some blocks, we further add overlapping blocks between adjacent 
rows of cropped images, with the size set to the average of cropped images in these rows.
For the horizontal direction, we also add overlapping blocks with the same size as those in the row.}
The cropping parameters $h_{\text{t}}$, $h_{\text{b}}$, $b_u$, $b_l$ can be set manually or automatically. Details are given in the supplementary document.

\subsection{Camera and Ground Plane Estimation}

\label{subsec:groundEstimation}
We use the ground plane as a guidance for three reasons:
1) it is a common element in large scenes, especially surveillance scenarios;
2) it is the main object interacting with people in the large scenes, reflecting the harmony between people and the scene;
3) it provides the important global information to the local cropped images.

To estimate the ground plane equation and the scene-level camera parameters,
the pose prior of people can be used 
\YL{for calibration}.
Note that the estimation of ground plane does not need too many people: more than ten people are enough as shown in the experiment (Sec. \ref{subsubsec:ablation}). 
Besides, our method can reconstruct people with various poses, but at the current stage, we only consider the standing or walking people who can be regarded as vertical lines on the ground plane. These people are automatically selected from the 2D keypoints detection obtained from RMPE \cite{fang2017rmpe}.
We use a pinhole camera model with a focal length $f\ (f = f_x = f_y)$ where the principal point $(c_x, c_y)$ of the camera is the image center.
We represent the ground equation as $N^T P_{g} + D = 0$, where $N=(x_n, y_n, z_n)$ is the ground normal with $\Vert N \Vert_2 = 1$, $P_{g} \in \mathbb{R}^3$ is the point on the ground plane and $D$ is \YL{a} constant term. 
For these standing people, 
we define the midpoints of their two ankle keypoints as $P_a \in \mathbb{R}^3$ and the midpoints of two shoulder keypoints as $P_s \in \mathbb{R}^3$ . 
The projections of $P_a$ and $P_s$ are $p_a = (u_a, v_a)$ and $p_s = (u_s, v_s)$, respectively. 
\YL{Following} perspective projection,
we have $z_a \times \bar{p}_a=K P_a$, where $\bar{p} = (u, v, 1)^T$ represents the homogeneous \YL{coordinates} of $p = (u, v)$,  $K$ is the intrinsic matrix of the scene-level camera and $z_a$ is the depth of $P_a$.
Similar to \cite{physicalDistance}, we assume that $P_a$ is on the ground plane, and the line from $P_a$ to $P_s$ is parallel to the ground normal. We also set a fixed height prior $h$.
Therefore, we have $N^T P_a + D = 0$ resulting in
\begin{equation}
    \label{eq:ground-transform}
    z_a=- \frac{D}{N^TK^{-1}\bar{p}_a},
\end{equation}
and $P_s$ can be approximated by $P_s' = P_a + h\times N$. Then, the projection $\bar{p}_{s}'$ is computed by
\begin{equation}
    \label{eq:reprojection}
    {z_{s}'}\times\bar{p}_{s}' = {z_{s}'}\times
    \begin{bmatrix}
    u_{s}'\\
    v_{s}'\\
    1
    \end{bmatrix}
    =K (z_a \times K^{-1} \bar{p}_a + h \times N).
\end{equation}
To solve the camera and ground plane parameters $K$, $N$, $D$, we adopt the following optimization loss:
\begin{equation}
\begin{aligned}
    \label{eq:re-projectionLoss}
    L_{\text{param}} = & \lambda_{\text{angle}} L_{\text{cos}}( p_{s}' - p_{a}, p_{s} - p_{a} ) \\ 
    & + \lambda_{\text{mod}} \frac {\left|\Vert p_{s}' - p_{a} \Vert_2 - \Vert p_{s} - p_{a} \Vert_2\right|}
                        {\Vert p_{s} - p_{a} \Vert_2},
\end{aligned}
\end{equation}
where $L_{\text{cos}}$ is the cosine distance, and $\lambda_{\text{angle}}$ and $\lambda_{\text{mod}}$ are the weights of the corresponding loss terms.
Finally, we translate $0.1$ meters along the ground normal direction to get the ground plane where people stand on.

\subsection{Crowd3DNet}

\label{subsec:crowd3dnet}

As shown in Fig.~\ref{fig:overview}, Crowd3DNet is a one-stage multi-head network based on the body-center-guided representation \cite{ROMP}.
Different from previous methods \cite{ROMP,BMP,BEV}, we define a new concept, \emph{Human-scene Virtual Interaction Point
(HVIP)}, and a progressive position transform (Sec. \ref{subsubsec:ground-guided}) to better infer the global 3D positions of people. Crowd3DNet outputs four maps including a body center heatmap, a torso center offset map, a 2D HVIP map and a SMPL parameters map.
The body center heatmap predicts the probability that each location is the center of a human body. 
If the body center heatmap gives positive responses, the network samples relevant parameters from other maps at the corresponding center locations to obtain 2D torso center offsets, 2D HVIPs and SMPL parameters of people.
With progressive position transform based on HVIP, Crowd3DNet combines the sampled parameters, the input of ground plane equation and scene-level camera parameters to infer accurate 3D positions of people, \YL{achieving multi-person reconstruction} in the large-scene camera system from the cropped images.


\subsubsection{Progressive Position Transform Based on HVIP}
\label{subsubsec:ground-guided}
\begin{figure}[!t]
    \centering
    \includegraphics[width=0.98\linewidth]{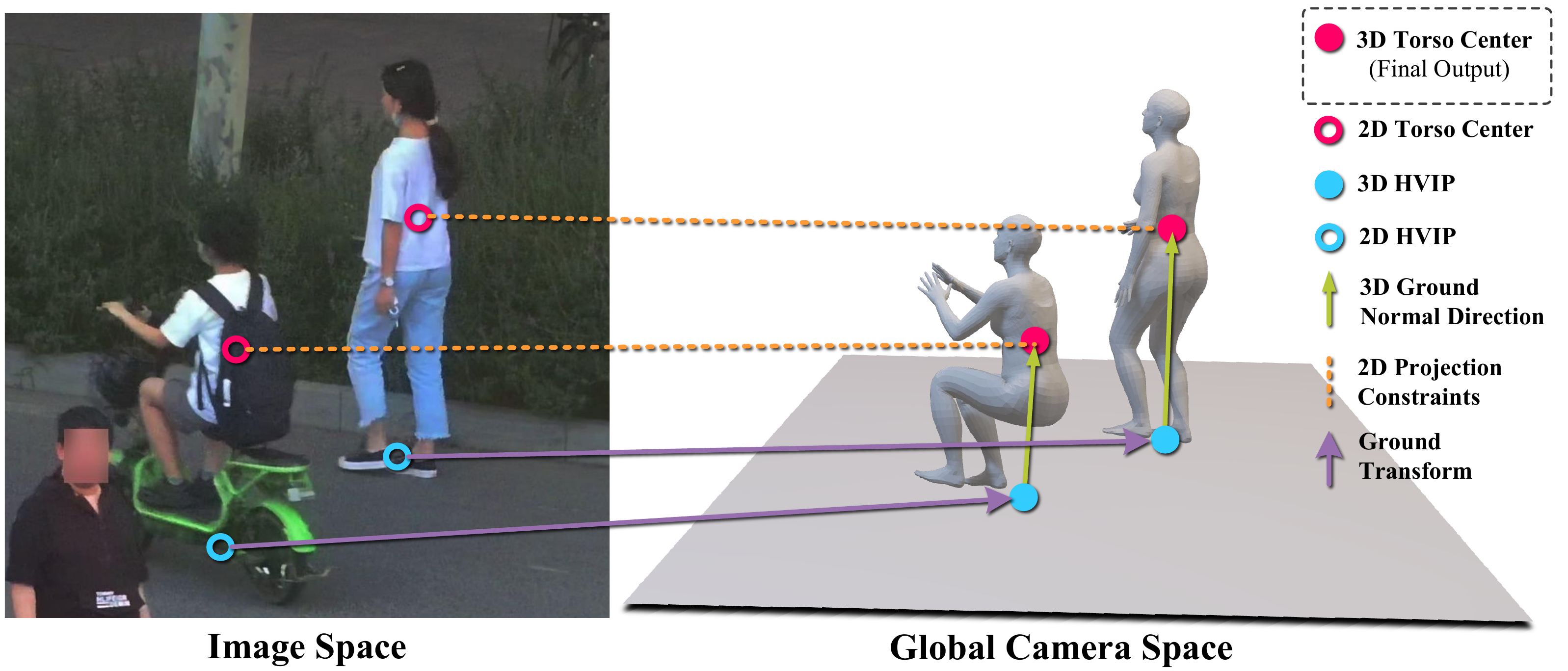}
    \vspace{-0.2cm}
    \caption{Progressive position transform based on HVIP.}
    \label{fig:hvip}
    \vspace{-0.4cm}
\end{figure}

We design the progressive position transform based on Human-scene Virtual Interaction Point (HVIP) to help infer the accurate 3D locations of persons in the large-scene camera system. 
The core idea is to infer the global 3D position from 2D image pixel points by HVIP and ground transform to avoid the depth ambiguity of estimating from a single view directly.
We define the HVIP which represents the projection point of a person's 3D torso center on the ground plane in the global camera space, marked as $P_{v}=(x_{v}, y_{v}, z_{v})$.
The torso center is a semantic point on human body, \ie, the center of two shoulder and two hip joints, 
represented as $P_{t}=(x_{t}, y_{t}, z_{t})$.
As show in Fig.~\ref{fig:hvip},
HVIP is a point on the ground plane, which can participate in ground transform directly to establish the mapping from image pixels to 3D points on the ground plane. 
HVIP binds a person body's semantic point but it is not on the human body itself. Therefore, different from previous method \cite{bodysize} that forces people's ankle joints \YL{to be} on the ground, \YL{which limits} the posture of people, HVIP \YL{is determined by} the 3D space position of people and can deal with people in various postures.
Because the line from $P_{t}$ to $P_{v}$ is perpendicular to the ground, we have $P_{t}=P_{v}+d\times N$, where $d$ represents the distance from $P_{t}$ to the ground plane.
We represent the projection points of $P_{v}$ and $P_{t}$ with $p_{v}=(u_{v}, v_{v})$ and $p_{t}=(u_{t}, v_{t})$, respectively.
Refer to Eq. \eqref{eq:ground-transform} and Eq. \eqref{eq:reprojection}, we deduce
\vspace{-0.2cm}
\begin{gather}
    \label{eq:tc}
    P_{v}=-\frac{D}{N^TK^{-1}\bar{p}_{v}} \times K^{-1}\bar{p}_{v}, \\
    d=\frac{f\times y_{v} - (v_{t}-c_y)\times z_{v}}{(v_{t}-c_y)\times z_{n} - f \times y_{n}}.
\end{gather}
Therefore, when the network predicts $p_{v}$ and $p_{t}$, $P_{t}$ can be uniquely determined.
Then, the predicted body mesh $M_{cam}$ in the global camera space follows $M_{cam}=M-P_{t-smpl}+P_{t}$, where $M \in \mathbb{R}^{6890 \times 3}$ and $P_{t-smpl} \in \mathbb{R}^{3}$ are the predicted vertices and torso center in SMPL \cite{smpl} space, respectively.
Finally, considering the cropping, our network takes the cropped image as input and predicts the local $p_{v-local}$ and $p_{t-local}$ on the cropped image. We have $p_{v}=p_{v-local}+t_{crop}$ and $p_{t}=p_{t-local}+t_{crop}$, where $t_{crop}$ represents the pixel \YL{coordinates} of the upper left corner of the cropped image.
Our progressive position transform with HVIP builds a mapping from some pixel points on \YL{the} cropped image to the global 3D location, which can simply and effectively predict the precise crowd positions.

\vspace{-0.3cm}
\subsubsection{Representations}
\vspace{-0.1cm}

\noindent\textbf{Input Parameters.}
The network cannot perceive the whole scene information only from the cropped image, hence we take the estimated ground and camera parameters as extra inputs. 
We define the camera input \YL{as} $(\frac{f}{W_s}, \frac{\hat{c}_{x} - c_x}{c}, \frac{\hat{c}_{y} - c_y}{c})$ which includes the information of FOV of scene and the principal point shift, where $W_s$, $(\hat{c}_{x}, \hat{c}_{y})$ and $c$ are the width of large-scene image, the image center of \YL{the} cropped image and the size of \YL{the} cropped image, respectively.

\noindent\textbf{Body Center Heatmap.}
The body center heatmap $\boldsymbol{C_m}$ represents the body center likelihood by a Gaussian kernel combining body scales, where $C_m \in \mathbb{R}^{1 \times H \times W}$ and $H=W=64$. We define the body center \YL{the} same as \cite{ROMP}.

\noindent\textbf{Torso Center Offset Map.}
Although we can directly define the body center as the 2D torso center, in practice, the body center heatmap tends to find a person's body salient point, especially when the person is occluded. Therefore, it is necessary to predict the human torso center separately. The torso center offset map $T_m \in \mathbb{R}^{2 \times H \times W}$ contains the offset between 2D torso center $p_{t-local}$ and body center.

\noindent\textbf{2D HVIP Map.} 
The goal of 2D HVIP map $H_m \in \mathbb{R}^{1 \times H \times W}$ is to obtain the 2D HVIP projection $p_{v-local}$ on the cropped image. 
The line from $P_{v}$ to $P_{t}$ is parallel to the ground normal, following the perspective theory, we have the projection points $p_{v}$, $p_{t}$ and the vanishing point of the ground normal $p_{vp}$ are collinear on image, where $p_{vp}=KN$. Therefore, we only need to estimate the 1D length from $p_t$ to $p_v$ to obtain 2D HVIPs.

\noindent\textbf{SMPL Map.} 
The SMPL map $S_m \in \mathbb{R}^{145 \times H \times W}$ includes the parameters of SMPL \cite{smpl} of people and a small 3D offset $\delta t$.
The SMPL parametric model can represent various shape and pose with a \YL{small} number of parameters.
It takes the pose parameters $\theta$ and the shape parameters $\beta$ as inputs and outputs a body mesh $M \in \mathbb{R}^{6890 \times 3}$. We adopt the 6D rotation representation \cite{zhou2019continuity} and drop the last two hand joints. \YL{Considering} the error of the dataset annotations, we predict \YL{an} offset $\delta t$ to further refine the position of people by $P_t=P_v+ d \times N + \delta t$.

\vspace{-0.2cm}
\subsubsection{Loss Function}
\label{subsec:loss functions}
\vspace{-0.1cm}
Crowd3DNet is supervised by the weighted sum of multiple loss \YL{terms} as follows:
\vspace{-0.1cm}
\begin{equation}
\begin{aligned}
    L=& \lambda_{\text{center}}L_{\text{center}}+
        \lambda_{\text{mesh}}L_{\text{mesh}}+
        \lambda_{\text{hvip}}L_{\text{hvip}} \\
        &+
        \lambda_{\text{tc}}L_{\text{tc}} +
        \lambda_{\text{root}}L_{\text{root}}+
        \lambda_{\text{gn}}L_{\text{gn}}+
        \lambda_{\text{out}}L_{\text{out}},
\end{aligned}
\end{equation}
\begin{equation}
\begin{aligned}
    L_{\text{mesh}}=& \lambda_{\text{pose}}L_{\text{pose}}+
        \lambda_{\text{shape}}L_{\text{shape}}+
        \lambda_{\text{j2D}}L_{\text{j2D}} \\  
        &+
        \lambda_{\text{j3D}}L_{\text{j3D}} +
        \lambda_{\text{paj3D}}L_{\text{paj3D}}+
        \lambda_{\text{gm}}L_{\text{gm}},
\end{aligned}
\end{equation}
where $L_{\text{center}}$ is 
the 2D focal loss \cite{lin2017focal}, and $L_{\text{mesh}}$ is the common SMPL related $L_2$ loss including pose parameter loss $L_{\text{pose}}$, shape parameter loss $L_{\text{shape}}$, 2D joint projection loss $L_{\text{j2D}}$, 3D joint loss $L_{\text{j3D}}$ and 3D joint loss after Procrustes alignment $L_{\text{paj3D}}$.
$L_{\text{hvip}}$, $L_{\text{tc}}$ and $L_{\text{root}}$ are all $L_2$ losses, which are used to supervise 2D HVIP projection, 2D torso center and absolute root position, respectively. 
$L_{\text{gn}} = L_{\text{cos}}(P_s - P_a, N)$ is a ground normal regularization term to enhance the interaction consistency between people and ground plane, where $P_s - P_a$ is the approximated craniocaudal direction of human.
We also use an out-of-bound loss to prevent people from penetrating the ground.
More concretely, we use $L_1$ loss to punish the point with the most serious penetration into the ground plane, and the out-of-bound loss is defined as
\vspace{-0.2cm}
\begin{equation}
    L_{\text{out}}=\left|\min(\{ \bar{v}_i \cdot G \mid \bar{v}_i \cdot G<0 \})\right|,
\end{equation}
where $v_i \in M_{\text{cam}}$ and $G=[N^T, D]^T$.

\subsection{Scene-specific Optimization and Merging}
\label{subsec:finetune}

To improve the harmony between reconstructed crowd and scene, and the generalization to various camera and ground plane parameters, we add a scene-specific optimization for a new scene at test time. 
Please note that the scene-specific optimization is performed only once for a camera-fixed scene, \ie, only one image of the scene is \YL{needed}.
Specifically, given a new scene at test time, we optimize a small set of weights in the head layer of Crowd3DNet with the ground normal and 2D poses 
\YL{estimated} in the camera and ground plane module. The optimization loss $L_{\text{opt}}$ is
\vspace{-0.1cm}
\begin{equation}
    L_{\text{opt}} = 
    \lambda_{\text{j2D}}L_{\text{j2D}} +
    \lambda_{\text{gm}}L_{\text{gm}} +
    \lambda_{\text{gn}} L_{\text{gn}} + 
    \lambda_{\text{out}}L_{\text{out}}.
\end{equation}

We finally remove \YL{duplicated} persons in the overlapped adjacent patches by merging. The merging operation retains the people farther away from the boundary of the overlapped region, which tends to keep more complete people to avoid truncation.

\vspace{-0.1cm} 
\section{Experiments}
\vspace{-0.1cm} 

\subsection{Large-scene Crowd Dataset}


To train and evaluate crowd reconstruction in a large scene, we contribute \emph{LargeCrowd}, which is a benchmark dataset with over 100K labeled humans in 733 gigapixel images ($19200\times6480$) of 9 different scenes (5 scenes for training and 4 scenes for \YL{testing}).
The images are extracted at a minimum interval of $3s$ from gigapixel streams which are captured by a ZoheTec JMC315 array camera.
We annotate the bounding boxes, 2D poses and 2D HVIPs of all the visible people in the images, with the maximum error less than 5 pixels for $95\%$ labels. We measure 3D landmarks in a world coordinate system and label the corresponding 2D points to solve the camera extrinsic matrix for each scene.
Then, we compute the homography matrix for each ground plane. The homography matrices with labeled ground segmentations and HVIPs provide the true physical positions of the persons. 


\begin{table}[!t]
    \caption{Comparison on \emph{LargeCrowd} dataset.}
    \vspace{-0.2cm}
    \centering
    \small
    \begin{tabular}{@{}c|cccc@{}}
    \toprule
    Method  &PPDS$\uparrow$  &PA-PPDS$\uparrow$    & PCOD$\uparrow$      &OKS$\uparrow$   \\
    \midrule
    SMAP\cite{smap}-Large  &58.60     & 60.07                 & 70.14        & 61.25         \\
    CRMH \cite{CRMH}-Large  &59.16      & 64.79               & 80.25         & 67.24         \\
    BEV \cite{BEV}-Large  & 74.21      & 75.05           & 87.31         & 66.15        \\
    \hline
    Crowd3D w/o HVIP            & 80.45       & 88.95       & 92.42          & 64.17        \\
    Crowd3D    &\textbf{81.53}       & \textbf{89.36}          & \textbf{92.63}        & \textbf{71.72}        \\
    \bottomrule
    \end{tabular}
\label{tab_large}
\vspace{-0.4cm}
\end{table}

\subsection{Implementation Details}
We use HRNet-32 \cite{cheng2020higherhrnet} as backbone, each head of which is composed of two ResNet \cite{he2016deep} blocks with batch normalization.
We resize input images to $512 \times 512$ with zero padding to keep the same aspect ratio. We also use the collision-aware representation of ROMP \cite{ROMP} to push apart close body centers. 
Our training process has two stages: 1) start by training the body center heatmap, torso center offset map and 2D HVIP map for 15 epochs to make sure that the subsequent learning about body mesh has a suitable initial position; 2) train the full model with all losses for 70 \YL{epochs}. We implement Crowd3DNet with PyTorch and adopt the Adam \cite{adam} as optimizer with 5e-5 learning rate. We train our model on \emph{LargeCrowd}, \emph{Agora} \cite{Agora}, \emph{MuCo-3DHP} \cite{MuCo3DHP} and a single person dataset \emph{Human3.6M} \cite{Human3.6M}.

\subsection{Evaluation Metrics}
We use the torso center as the location of people and evaluate the location distribution of crowd by the distances between people.
We define a metric called pair-wise percentual distance similarity (PPDS) as
\vspace{-0.2cm}
\begin{gather}
    \textit{PPDS}=
    \frac{\sum_{k=1}^{n-1} \sum_{i=k+1}^{n} 1- \min{(d_{ik}, 1)}}{C_n^{2}}, \\
    d_{ik}=\left | \frac{ \Vert E_k - E_i \Vert -\Vert G_k - G_i \Vert }{\Vert G_k - G_i \Vert}\right |,
\end{gather}
where $n$ is the number of people in the image, and $E_i$ and $G_i$ represent the estimated and ground-truth locations of the $i$-th person, respectively.
To evaluate the relative crowd distribution, we also define the procrustes-aligned pair-wise percentual distance similarity (PA-PPDS) which aligns the reconstructed crowd and the ground truth by Procrustes alignment to exclude the influence of scale and rotation.
Due to the lack of 3D pose annotations, we use the object keypoint similarity (OKS) \cite{coco} to evaluate the 2D poses. The percentage of correct ordinal depth (PCOD) \cite{smap} is used to evaluate the ordinal depth relations between all pairs of people in the image. 

\begin{figure*}[!h]
    \centering
    \includegraphics[width=0.8\linewidth]{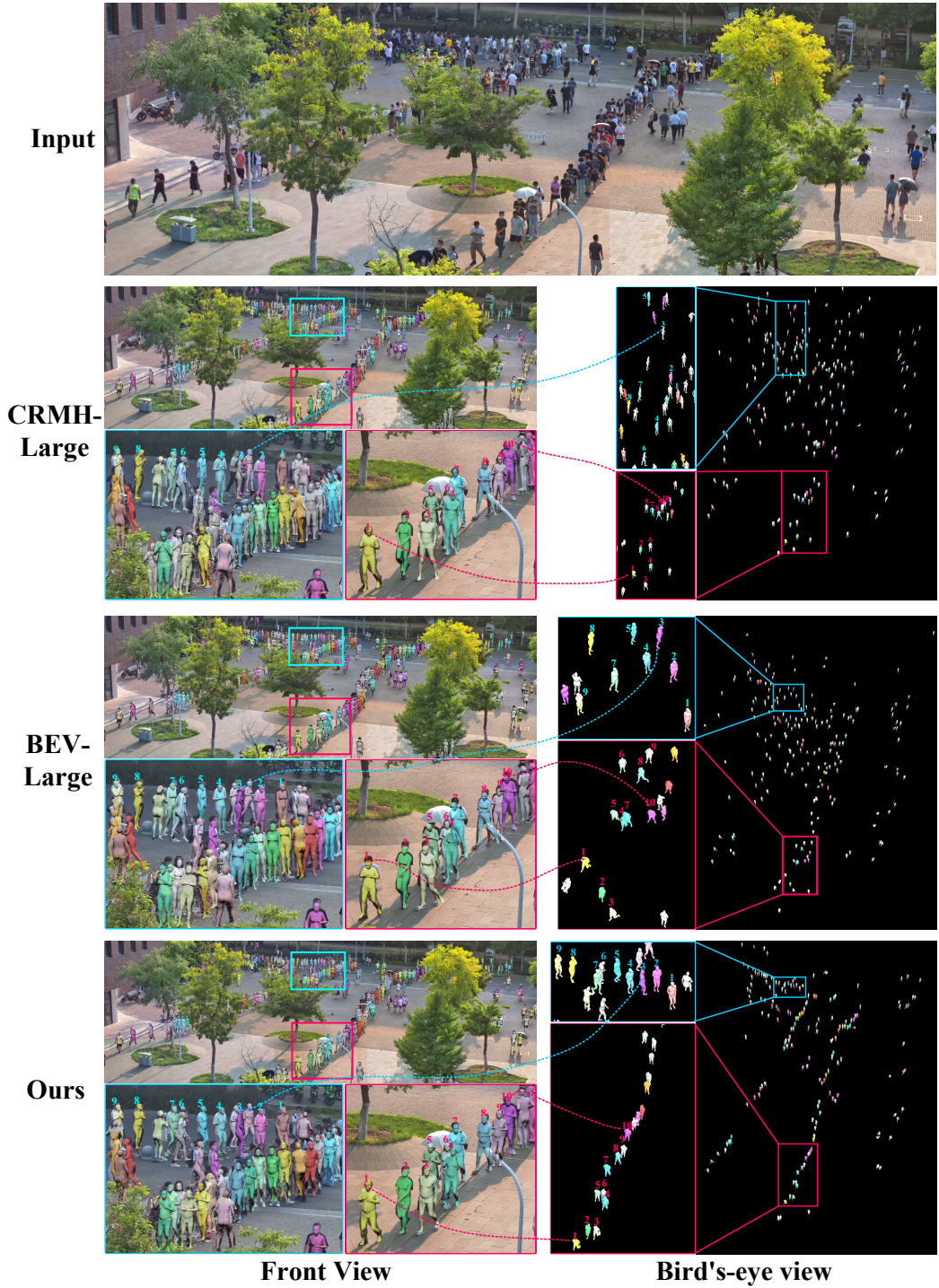}
    \vspace{-0.2cm}
    \caption{Qualitative results on \emph{LargeCrowd}. The same color or number corresponds to the same person, and gray indicates that the person is not matched.}
    \label{fig:comparision}
      \vspace{-0.4cm}  
\end{figure*}

\vspace{-0.1cm}
\subsection{Comparison}
\vspace{-0.1cm}
Because no existing methods can directly handle large-scene images with hundreds of people, we compare our method with three baselines that are modified \YL{from} the state-of-the-art methods: SMAP \cite{smap}, CRMH \cite{CRMH}, and BEV \cite{BEV}. We denote these baselines as SMAP-Large, CRMH-Large and BEV-Large. 
Specifically, we first use our adaptive human-centric cropping to obtain the hierarchical cropped images as their inputs and infer the respective reconstructed results on the cropped images. To obtain the global reconstruction results for these methods, we provide the scene-camera intrinsics estimated by our method to them.
For CRMH \cite{CRMH} which predicts human \YL{bodies} in bounding \YL{boxes} by \YL{a} weak perspective camera model, we use its transform from bounding box position to depth of full image to infer the predicted locations in the global camera space.
Both SMAP \cite{smap} and BEV \cite{BEV} infer the locations through perspective camera models. Following previous method \cite{albiero2021img2pose}, we scale the depths of their results according to the focal length of \YL{the} scene. Please refer to  supplementary material for more details. For fair comparison, we fine-tune all the compared methods on \emph{LargeCrowd}.
Table \ref{tab_large} gives the quantitative results. 
Our method outperforms other approaches in terms of all the metrics. Especially, the obvious advantage in PPDS, PA-PPDS and PCOD shows that our method can  predict accurate crowd location distribution, including physical distances and relative arrangements.
Fig.~\ref{fig:comparision} shows qualitative comparison results. 
The complete bird's-eye view on the right shows that our predicted crowd distribution is consistent with the input image, while the compared methods are not consistent.
Taking the persons \YL{labeled} with numbers for example, only our method recovers correct relative positions. 
The reconstructed people by the existing methods independently inferred from the cropped images are inconsistent in the global large-scene camera space.
Besides, although these methods show reasonable projection results, the wrong global positions mean that their predicted 3D human bodies have wrong scales. We also provide comparison results on public small-scene datasets in supplementary material.


\subsection{Ablation Study}
\label{subsubsec:ablation}
\begin{figure}[!t]
    \centering
    \includegraphics[width=0.8\linewidth]{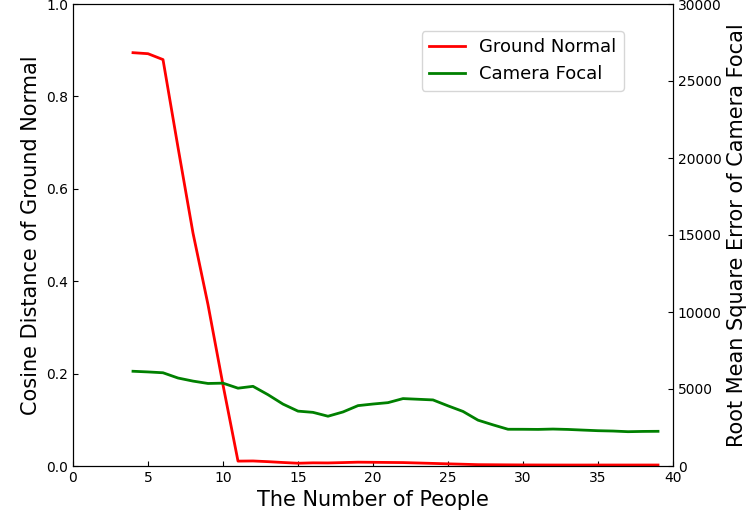}
    \vspace{-0.2cm}
    \caption{The impact of the number of people on the camera and ground plane estimation.}
    \label{fig:personNum}
    \vspace{-0.4cm}
\end{figure}

\noindent\textbf{Impact of the Number of People on the Estimated Ground and Camera.} 
We explore the impact of the number of people on estimating ground and camera parameters by controlling the number of people used in optimization, and the newly added people are randomly selected.
The metrics include a cosine distance for ground normal and a root mean square error for focal length. We calibrate the camera and obtain the ground-truth focal length (about 27000).
As shown in Fig.~\ref{fig:personNum}, more than ten people are enough for estimating ground normal, which is common in real-world large-scale scenes, especially surveillance scenarios with hundreds of people. The focal length is not sensitive to the number of people.

\noindent\textbf{Progressive Position Transform Based on HVIP.} 
Our progressive position transform based on HVIP effectively helps the network to predict accurate global 3D positions of people. 
To verify this, we compare our full model with a variant of Crowd3DNet, Crowd3D w/o HVIP, \wh{which predicts 2D ankle joints and adopts the midpoint of ankle joints to participate in ground transform without using HVIP. }
The result is shown in Table~\ref{tab_large}.
Benefiting from HVIP, which makes use of the ground plane without restricting human posture, Crowd3DNet has obvious advantages on OKS.

\noindent\textbf{Adaptive Human-centric Cropping.}
To verify the adaptive human-centric cropping scheme, we denote a metric called cropping score, which counts the ratio of people with the appropriate scale after cropping. 
For a person with appropriate scale, we set the ratio of his height to the corresponding cropped image within [0.3, 0.8], and he is not truncated.
The comparison result on \emph{LargeCrowd} between adaptive human-centric cropping and uniform cropping is 0.923 \emph{vs.} 0.806, which demonstrates the effectiveness of our adaptive human-centric cropping scheme.

\section{Conclusion and Discussion}
\label{Sec:Conclusion}

\noindent\textbf{Conclusion.}
We propose Crowd3D to reconstruct hundreds of people with global consistency from a single RGB large-scene image. Our method is a joint local and global inference framework which converts the complex crowd localization into pixel localization by our defined HVIP concept and the parameters of pre-estimated scene-level camera and ground plane.
Our adaptive human-centric cropping scheme and progressive position transform based on HVIP solve the challenges of large number of people, large variations in human scale and complex spatial distribution in large scenes.
We also contribute a large-scene dataset called \emph{LargeCrowd} to help train and evaluate crowd reconstruction in large scenes with hundreds of people.
Experimental results demonstrate that our method can achieve globally consistent crowd reconstruction in large scenes.

\noindent\textbf{Limitations and Future Work.}
We focus on outdoor real-world large-scale scenes which contain one or several ground planes. 
Our method may be easily extended to multi-ground scenes by using the existing image-based ground plane segmentation methods or manual segmentation, which is taken as our future work.
Although our \YL{Crowd3D} shows effective crowd reconstruction in a global camera space, there are still some cases that we cannot solve well, \eg, the people in complex ground conditions and the persons with complicated postures or severe occlusions.
In future work, we will focus on a wider range of large-scale scenes with complex ground and crowd environments.

\noindent\textbf{Acknowledgments.}
This work was supported in part by the National Natural Science Foundation of China (62122058, 62125106, 61860206003 and 62171317).

{\small
\bibliographystyle{ieee_fullname}
\bibliography{egbib}
}

\end{document}


\title{Supplementary Document for\\
Crowd3D: Towards Hundreds of People Reconstruction from a Single Image}

\author{
    Hao Wen$^{1, \dagger}$, Jing Huang$^{1, \dagger}$, Huili Cui$^{1}$, Haozhe Lin$^{2}$, Yu-Kun Lai$^{3}$, Lu Fang$^{2}$, Kun Li$^{1,*}$\\
    $^{1}$Tianjin University, China \enspace  $^{2}$Tsinghua University, China \enspace $^{3}$Cardiff University, United Kingdom\\
    {\tt\small \{wenhao, hj00, huilicui\_1, lik\}@tju.edu.cn,
    \{linhz, fanglu\}@tsinghua.edu.cn,}\\
    {\tt\small LaiY4@cardiff.ac.uk}
}

\maketitle
\thispagestyle{empty}

\let\thefootnote\relax\footnotetext{$\dagger$ Equal contribution.}
\let\thefootnote\relax\footnotetext{* Corresponding author.}

In this document, we provide more paper details, including:
\begin{itemize}
  \item How to automatically set cropping parameters;
  \item How to obtain the results of other methods for large-scene images;
  \item Quantitative results on small scenes;
  \item \wh{More ablation studies; }
  \item Qualitative results on \emph{PANDA}.
\end{itemize}

We also provide a demo video along with this document.

\section{Cropping Parameter Settings }
Our adaptive human-centric cropping scheme crops a large-scene image into patches with hierarchical sizes which ensures that the height ratio between the person and the corresponding image is as consistent as possible among different cropped images. This is beneficial for the subsequent inference of human poses, shapes and locations.
To obtain the cropping parameters automatically, 
we first crop the images with sliding windows of different scales to ensure that each object is detectable and then use a state-of-the-art pose estimation method \cite{fang2017rmpe} to obtain pre-detection results.
We obtain the 2D poses of persons at the top and the bottom of the large-scene image by sorting $y$ coordinate of centers of detected poses, and set the heights $h_{t}$ and $h_{b}$ of persons at the top and the bottom by respective pose results.
We assume that the persons at the top and the bottom are in the center of their respective cropped images, hence the upper and lower bounds of the image area to be processed are set as $b_u=c_{y_{t}}-h_{t}$ and $b_l=c_{y_{b}}+h_{b}$, where $c_{y_{t}}$ and $c_{y_{b}}$ are the $y$ coordinate of pose centers of corresponding persons.
We also use some tricks to improve the reliability of the obtained cropping parameters, including detecting multiple images of a scene, requiring the size of the person at the bottom to be larger than the vast majority of the detected people, \etc.
In general, automatically obtaining the cropping parameters is sensitive to false detections and has a high time cost. Therefore, we recommend setting the cropping parameters $h_t$, $h_b$, $b_u$ and $b_l$ manually, which is easy and has a low time cost.

\section{Running Other Methods on Large-scene Images}

Existing methods cannot directly process large-scene images.
To obtain the crowd reconstruction results on large-scene images using these methods, we apply our adaptive human-centric cropping to obtain the hierarchical cropped images as their inputs and provide the estimated scene-level focal length $f_s$ of our method.

For CRMH \cite{CRMH}, 
we use its own camera coordinate transformation to convert the camera parameters corresponding to human bounding boxes of cropped images to the global scene-level depths.
We restore the bounding boxes of cropped images to the pixel coordinates of large-scene images by upper-left coordinates of the cropped images.
We represent the camera parameters corresponding to the bounding box $B_{i}=\left[x_{\min }, y_{\min }, x_{\max }, y_{\max }\right]$ of the $i $-th person with $ \boldsymbol{\pi}_{i}=\left\{s_{i}, x_{i}, y_{i}\right\} $, and define the center and size of  $B_i$ as $ c_{i}=\left[\left(x_{\min }+x_{\max }\right) / 2,\left(y_{\min }+y_{\max }\right) / 2\right] $ and $ \alpha_{i}=\max \left(x_{\max }-x_{\min }, y_{\max }-y_{\min }\right) $, respectively. 
Given these parameters, the global depth of the $i$-th person is calculated as
\begin{equation}
d_{i}=\frac{2 f_s}{s_{i} \alpha_{i}}.
\end{equation}
Then, the global translation of the $i$-th person can be obtained by
\begin{equation}
T_{i\_global}=\left[\begin{array}{c}
d_{i}\left(x_{i} \alpha_{i}+c_{i, x}-w_s / 2\right) / f_s \\
d_{i}\left(y_{i} \alpha_{i}+c_{i, y}-h_s / 2\right) / f_s \\
d_{i}
\end{array}\right],
\end{equation}
where $ w_s $ and $ h_s $ are the width and height of the large-scene image, respectively.

For SMAP \cite{smap} and BEV \cite{BEV}, we use the method in \cite{albiero2021img2pose} to implement the local-to-global depth conversion.
For the cropped image with width $ w_{c} $ and height $ h_c $,
we set the focal length $f_c$ of local camera system for SMAP and BEV by respective model settings. We have $f_c=w_c$ for SMAP and $f_c=f_{b}\times w_c / s_{b}$ for BEV, where $f_{b}=443.4$ and $s_b=512$ are the focal length and the input image size of BEV model. 
We represent the torso center of the $i$-th person in the local camera coordinate system with $T_{i}=\left\{X_{i},Y_{i},Z_{i}\right\} $, and the depth in the global coordinate system is calculated as $ Z_{i\_global}=Z_{i} \times f_s / f_c $. 
We also restore the 2D pose pixel coordinates of cropped images to large-scene image by the positions of cropped images.
With the 2D projection $ \left\{x_{i},y_{i}\right\}$ of torso center on the large-scene image, the global depth $ Z_{i\_global} $ and our scene-level camera parameters $ K $, the global position of the $ i $-th person can be calculated as
\begin{equation}
T_{i\_global}=Z_{i\_global} K^{-1}\left[\begin{array}{c}
x_{i} \\
y_{i} \\
1
\end{array}\right].
\end{equation}


\section{Quantitative Results on Small Scenes}
\wh{We evaluate our method on the small-scene datasets \emph{Panoptic}\cite{Panoptic} and \emph{MuPoTS}\cite{MuCo3DHP}, compared with state-of-the-art methods.
We directly use small-scene images as inputs to our Crowd3DNet.
The ground plane equations that we use are estimated by combining the people of the same scene at different frames since there are only 2-6 people in a small scene image.
For \emph{Panoptic}, we use MPJPE (mean per joint position error), root error (RtError) and percentage of correct ordinal depth (PCOD) to evaluate the 3D poses and locations of the reconstructed people. We do not test on \emph{Mafia} because the related 3D annotations cannot be obtained from the official website. We use the SMAP \cite{smap} model provided by the authors that is not trained on \emph{Panoptic} for fair comparison. 
As shown in Table \ref{tab_small}, our method achieves the best results,  validating the performance of our method on both position inference and pose estimation for small scenes.
For \emph{MuPoTS}, we follow the protocol of \cite{3DMPPE}.
The results in Table \ref{tab:mupots} also demonstrate the effectiveness of our method.}

\begin{table}[h]
    \centering
    \small
    \caption{Comparisons to the state-of-the-art methods on a small-scene dataset \emph{Panoptic}. }
    \begin{tabular}{@{}c|ccccc@{}}
    \toprule
    \multicolumn{1}{l|}{}    & Method          & \emph{Haggling}        & \emph{Ultim}         & \emph{Pizza}          & Mean\\
    \midrule
    \multirow{8}{*}{MPJPE $\downarrow$}   
        & SMAP \cite{smap}             & 128.5              & 141.2          & 236.4          & 168.7\\
        & CRMH \cite{CRMH}            & 129.6           & 153.0          & 156.7          & 146.4\\
        
        & BMP \cite{BMP}             & 120.4           & 140.9          & 147.5          & 136.3\\
        & ROMP \cite{ROMP}           & 110.8          & 141.6          & 137.6         & 130.0\\
        & BEV \cite{BEV}            & 100.9      & 132.4  & 139.6 &  124.3\\
        
        & Crowd3D & \textbf{100.1}             & \textbf{125.7}          & \textbf{134.1}          & \textbf{120.0}\\
        \midrule
    \multirow{5}{*}{RtError $\downarrow$ } 
        & SMAP \cite{smap}            & 432.8             & 529.8        & 1297.6 & 753.4\\
        & CRMH \cite{CRMH}            & 2384.5            & 2301.0         & 2418.7         & 2368.1\\
        & BEV \cite{BEV}            & 786.1           & 683.0          & 763.2          & 744.1\\
        & Crowd3D                   & \textbf{274.8}    & \textbf{295.3}  & \textbf{542.1}  & \textbf{370.7}\\
    \midrule
    \multirow{5}{*}{PCOD $\uparrow$ }    
        & SMAP \cite{smap}            & 83.5               & 93.3           & 76.0           & 84.3\\
        & CRMH \cite{CRMH}            & 89.5            & 93.2           & 74.8           & 85.8\\
        & BEV \cite{BEV}             & 89.3              & 98.3  &\textbf{ 94.7 }          & 94.1\\
        & Crowd3D & \textbf{90.4}      & \textbf{99.5}          & 92.8 & \textbf{94.2}  \\
    \bottomrule
    \end{tabular}
    \label{tab_small}
\end{table}

\begin{table}[tp]
    \centering

    \footnotesize
    \caption{Comparison on \emph{MuPoTS} dataset.}

    \begin{tabular}{@{}c|cc@{}}
        \toprule
        Method  &ALL$\uparrow$  &Matched$\uparrow$  \\
        \midrule
        CRMH  &69.1     & 72.2  \\
        ROMP &69.9     &74.6\\
        BEV  &70.2     & 75.2  \\
        Ours  &\textbf{70.9}  & \textbf{75.4} \\
        \bottomrule
    \end{tabular}
    \label{tab:mupots}

\end{table}



    

\begin{table}[!t]

    \caption{Quantitative ablation study on \emph{LargeCrowd} dataset.}

    \centering
    \footnotesize
    \begin{tabular}{@{}c|cccc@{}}
    \toprule
    Method  &PPDS$\uparrow$  &PA-PPDS$\uparrow$    & PCOD$\uparrow$      &OKS$\uparrow$   \\
    \midrule
    
    w/o AC, w/o GT-GC   & 81.5          & 89.1       & 92.3          & 69.5      \\
    w/ AC, w/o GT-GC    &81.5       & 89.4       & 92.6       & 71.7        \\
    w/ AC, w/ GT-GC &92.4   & 92.4    & 93.5 & 73.9\\
    \bottomrule
    \end{tabular}
\label{tab_ablation}
\end{table}

\section{Ablation Studies}
\wh{Besides ablation studies on the HVIP and the cropping score given in the paper, Tab. \ref{tab_ablation} gives additional ablations of some crucial modules (AC: adaptive cropping, GT-GC: ground-truth ground and camera parameters).
The first two lines show the positive impact of adaptive human-centric cropping scheme on crowd reconstruction.
The last line shows that better ground estimation can further improve the performance of our method.}

\section{Qualitative Results on PANDA}
\begin{figure*}[!htp]
    \centering
    \includegraphics[width=0.95\linewidth]{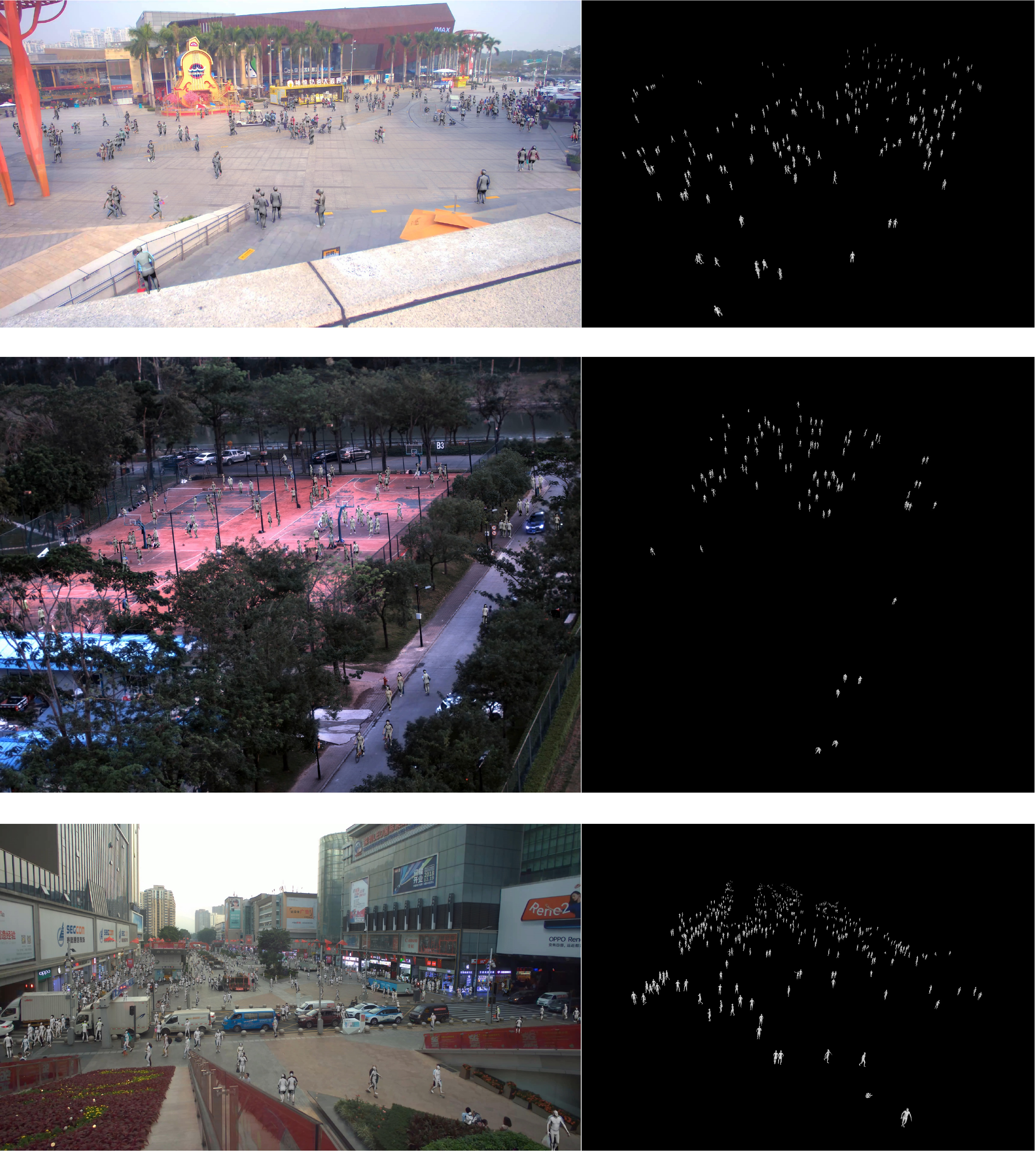}
    \caption{Qualitative results on \emph{PANDA}.}
    \label{fig:panda}
\end{figure*}

In Fig. \ref{fig:panda}, we present qualitative results on \emph{PANDA}\cite{PANDA}.
\emph{PANDA} is a human-centric large-scene video dataset which has gigapixel-level resolutions and contains hundreds of people.
Please note that the gigapixel camera used in \emph{PANDA} is different from the camera in \emph{LargeCrowd}, and the image resolutions of the two datasets are also different. The consistent reconstructed results show the generalization ability of our method.


{\small
\bibliographystyle{ieee_fullname}
\bibliography{egbib}
}